\documentclass[final,5p,times,twocolumn]{elsarticle}
\usepackage{framed,multirow}

\usepackage{amssymb}
\usepackage{amsmath}
\usepackage{latexsym}
\usepackage{latexsym}
\usepackage{pifont}
\usepackage{subcaption}
\usepackage{hyperref}

\usepackage{url}
\usepackage[dvipsnames]{xcolor}

\begin{document}

\begin{frontmatter}

\title{Learn to cycle: Time-consistent feature discovery for action recognition}
\author{Alexandros Stergiou} 
%\cortext[cor1]{Corresponding author:}
\ead{a.g.stergiou@uu.nl}
\author{Ronald Poppe}

\address{Utrecht University, Princetonplein 5, 3584 CC Utrecht, The Netherlands}

\begin{abstract}
Generalizing over temporal variations is a prerequisite for effective action recognition in videos. Despite significant advances in deep neural networks, it remains a challenge to focus on short-term discriminative motions in relation to the overall performance of an action. We address this challenge by allowing some flexibility in discovering relevant spatio-temporal features. We introduce Squeeze and Recursion Temporal Gates (SRTG), an approach that favors inputs with similar activations with potential temporal variations. We implement this idea with a novel CNN block that uses an LSTM to encapsulate feature dynamics, in conjunction with a temporal gate that is responsible for evaluating the consistency of the discovered dynamics and the modeled features. We show consistent improvement when using SRTG blocks, with only a minimal increase in the number of GFLOPs. On Kinetics-700, we perform on par with current state-of-the-art models, and outperform these on HACS, Moments in Time, UCF-101 and HMDB-51.\footnotemark
\end{abstract}

\end{frontmatter}
%\linenumbers

\footnotetext{The code for this project can be found at: \url{https://git.io/JfuPi}}

%% main text
\section{Introduction}
\label{introduction}

% Introduction to progress made in action recognition
Action recognition in videos is an active field of research. A major challenge that is addressed comes from dealing with the vast variation in the temporal display of the action \citep{herath2017going,stergiou2019analyzing}. In deep neural networks, temporal motion has primarily been modeled either through the inclusion of optical flow as a separate input stream \cite{simonyan2014two} or using 3D convolutions \cite{ji20133d}. The latter have shown consistent improvements in state-of-the-art models \citep{carreira2017quo,chen2018multifiber,feichtenhofer2019slowfast,feichtenhofer2020x3d}.

% Structure of 3D CNN blocks and temporal locality
3D convolution kernels in convolutional neural networks (3D-CNNs) take into account fixed-sized temporal regions. Kernels in early layers have small receptive fields that primarily focus on simple patterns such as texture and linear movement. Later layers have significantly greater receptive fields that are capable of modeling complex spatio-temporal patterns. Through this hierarchical dependency, the relations between discriminative short-term motions within the larger motion patterns are only established in the very last network layers. Consequently, when training a 3D-CNN, the learned features might include incidental correlations instead of consistent temporal patterns. Thus, there appears to be room for improvement in the discovery of discriminative spatio-temporal features.

\begin{figure}[!t]
\centering
\includegraphics[width=\linewidth]{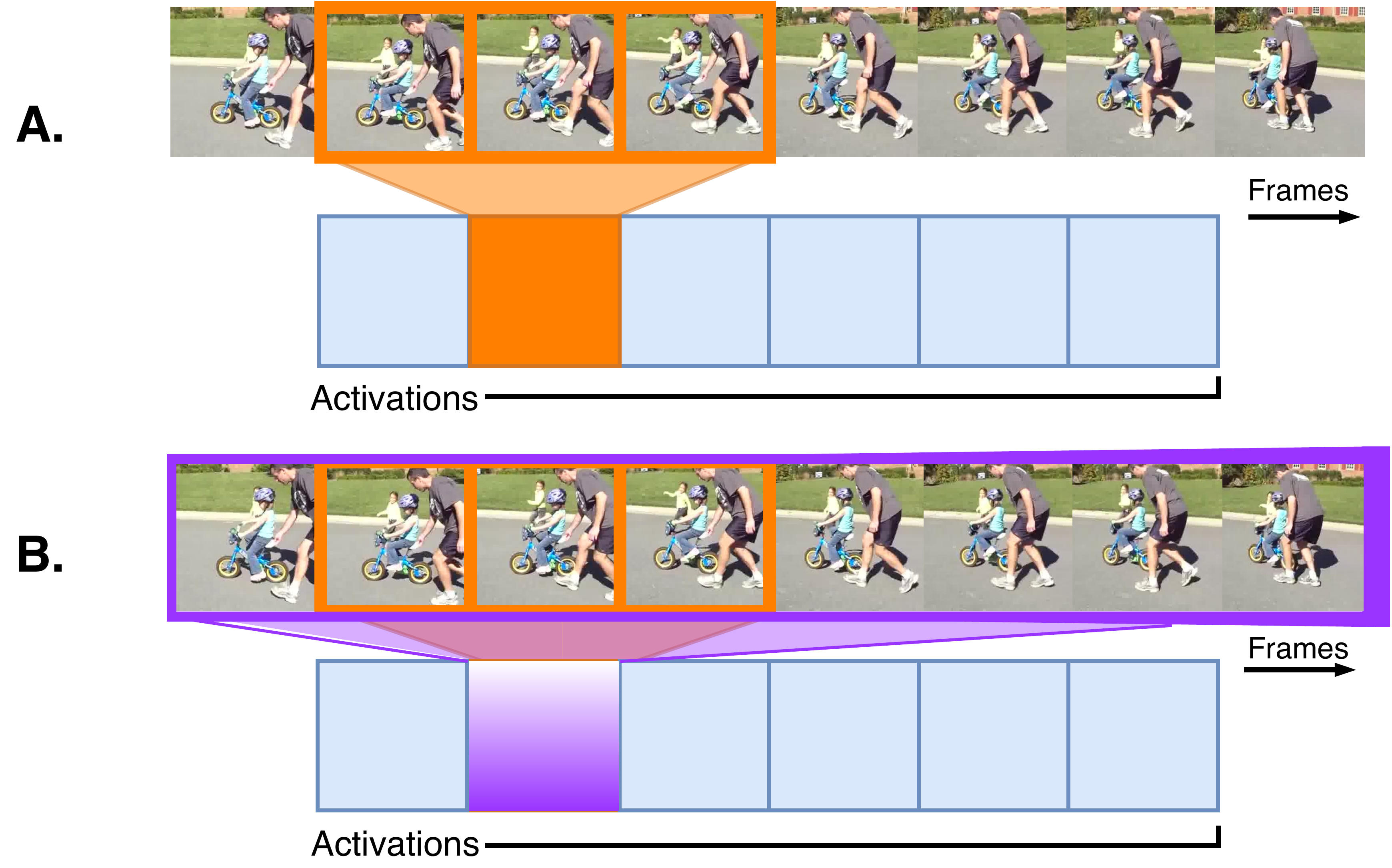}
\caption{A. Original 3D convolution block. Activation maps consider a fixed-size temporal window. Features are specific to the local neighborhood. B. SRTG convolution block. Activation maps take global time information into account.}
\label{fig:SRTG_PoC}
\end{figure}

\begin{figure*}
\centering
\includegraphics[width=0.9\textwidth]{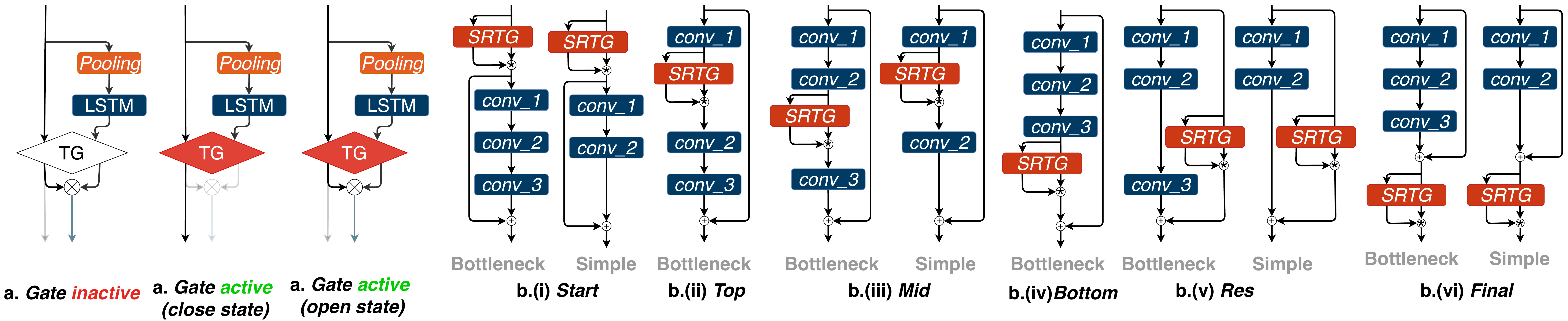}
\caption{(a) SRTG gate states. The gates can be \textcolor{red}{inactive} or \textcolor{green}{active}. When inactive, main stream and LSTM stream are fused. When active, the output is determined by the Temporal Gate and is either the fused result (open gate) or only the main stream (close state). (b) SRTG configuration options described in Section~\ref{sec:Variants}. Similar to Residual Networks, we distinguish between \textit{Simple} blocks with two conv operations and \textit{Bottleneck} blocks with three conv operations.}
\label{fig:SR_pipeline}
\end{figure*}

% Overview of the proposed method (SRTG)
To improve this discovery process, we propose a method named \textit{Squeeze and Recursion Temporal Gates} (SRTG) which aims at extracting features that are consistent in the temporal domain. Instead of relying on a fixed-size window, our approach relates specific short-term activations to the overall motion in the video, as shown in Figure~\ref{fig:SRTG_PoC}. We introduce a novel block that uses an LSTM \citep{hochreiter1997long}) to encapsulate feature dynamics, and a temporal gate to decide whether these discovered dynamics are consistent with the modeled features. The novel block can be used as at various places in a wide range of CNN architectures, with minimal computational overhead.

% Key contributions
Our contributions are as follows:
\vspace{-1mm}
\begin{itemize}
    \item We implement a novel block, Squeeze and Recursion Temporal Gates (SRTG), that favors inputs that are temporally consistent with the modeled features.
    
    \item The SRTG block can be used in a wide range of 3D-CNNs, including those with residual connections, with minimal computational overhead ($\sim$0.15\% of model GFLOPs).

    \item We demonstrate state-of-the-art performance on five action recognition datasets when SRTG blocks are used. Networks with SRTG consistently outperform their vanilla counterparts, independent of the network depth, the convolution block type and dataset.
    
\end{itemize}

% Paper Structure overview
We discuss the advancements in the modeling of time in action recognition in Section~\ref{sec:related}. A detailed description of the main methodology is provided in Section~\ref{sec:methodology}. Experimental setup and results are presented in Section~\ref{sec:results} and we conclude in Section~\ref{sec:conclusions}.

\section{Related Work}
\label{sec:related}
We discuss how temporal information is represented in CNNs, in particular using 3D convolutions.

% 3D convolutions.
\textbf{Time representation in CNNs.} Apart from the hand-coded calculation of optical flow \citep{simonyan2014two}, the predominant method for representing spatio-temporal information in CNNs is the use of 3D convolutions. These convolutions process motion information jointly with spatial information \citep{ji20133d}. Because the spatial and temporal dimensions of videos are strongly connected, this has led to great improvements especially for deeper 3D-CNN models \citep{carreira2017quo,hara2018can}. Recent work additionally targets the efficient incorporation of temporal information at different time scales through the use of separate pathways \citep{chen2018multifiber,feichtenhofer2019slowfast}.

% 3D variants and temporal receptive field size.
\textbf{3D convolution variants.} A large body of work has focused on reducing the computational requirements of 3D convolutions. Most of these attempts are targeted towards the decoupling of temporal information, for example as \textit{pseudo} and (2+1)D 3D convolutions \citep{qiu2017learning,tran2018closer}. Others have proposed a decoupling of horizontal and vertical motions \citep{stergiou2019FAST}.

% Improvements based on attention and information fusion in activations.
\textbf{Information fusion of spatio-temporal activations.} \textit{Squeeze and Excitation} \citep{hu2018squeeze}, \textit{Gather and Excite} \citep{hu2018gather} and \textit{Point-wise Spatial Attention} \citep{zhao2018psanet} consider self-attention in convolutional blocks for image-based input. In the video domain, self-attention has been implemented by \citet{long2018attention} using clustering, to integrate local patterns with different attention units. Others have studied the use of non-local operations that capture long-range temporal dependencies through different distances \citep{wang2018non}. \citet{wang2018appearance} proposed to filter feature responses with activations decoupled to branches for appearance and spatial relations. \citet{qiu2019learning} have extended the idea of creating separate pathways for general features that can be updated through network block activations.

% Need for Inclusion of feature temporal variations.
While these methods have shown increased generalization performance, they do not address the discovery of local spatio-temporal features across large time sequences. As activations are constrained by the spatio-temporal locality of their receptive fields, they are not allowed to effectively consider extended temporal variations of actions based on their general motion and time of execution. Instead of attempting to map the locality of features to each of the frame-wise activations, our work combines the locally-learned spatio-temporal features with their temporal variations across the duration of the video sequence.

\section{Squeeze and Recursion Temporal Gates}
\label{sec:methodology}

% Overview of section and general comments on the math to follow.
In this section, we introduce Squeeze and Recursion Temporal Gates (SRTG) blocks, and the possible configurations for their use in CNNs. We will denote layer input $a$ as a stack of $T$ frames $a_{\: (C \: \times \: T \: \times \: H \: \times \: W)}$ with $C$ the number of channels, $T$ the number of frames, and $H$ and $W$ the spatial dimensions of the video. The backbone blocks that SRTG are applied to also include residual connections where the final accumulated activations are the sum of the previous block activations ($a^{[l-1]}$) and the current computed features ($z^{[l]}$) denoted as $a^{[l]} = z^{[l]} + a^{[l-1]}$, with block index $l$.

\subsection{Squeeze and Recursion}
\label{sec:SR}

% Activations pre-processing blocks
Squeeze and Recursion blocks can be built on top of any spatio-temporal activation map $a^{[l]} = g(z^{[l]})$ for any activation function $g()$ applied to a volume of features $z^{[l]}$, shown in Figure~\ref{fig:SR_pipeline}(a). This process is similar to Squeeze and Excitation~\citep{hu2018squeeze}. For each block, the activation maps are sub-sampled in both spatial dimensions to create a vectorized representation of the volume's features across time. Each element in the vector contains the intensity values of a frame squeezed, so to say, in a single average value. This process encapsulates the average temporal attention through the discovered features.

%as in Equation~\ref{eq:global_spatial_pool}
%\vspace{-0.5mm}
%\begin{equation}
%    \label{eq:global_spatial_pool}
%    pool(a^{[l]}) = a^{[l]}_{ \; (C \; x \; T \; x \; \overline{H} \; x \; \overline{W})} \; ,where \; %|\overline{H}|=1 \: and \: |\overline{W}|=1
%\end{equation}

\begin{figure}[!htb]
\centering
\includegraphics[width=\linewidth]{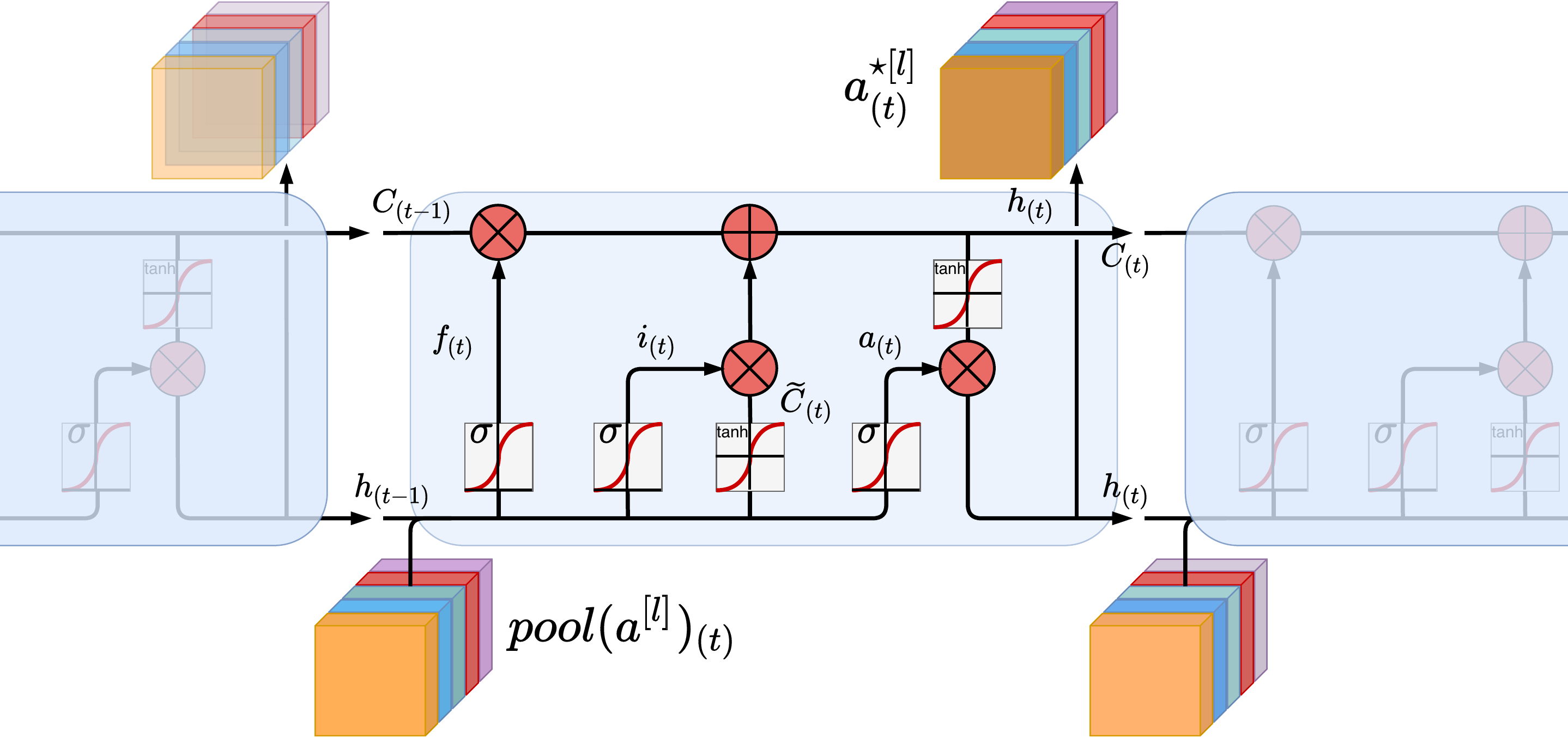}
\caption{Overview of the LSTM-chained cells used for the discovery of globally informative local features. Each input corresponds to a temporal activation map and produces a feature vector of the same size as the input.}
\label{fig:LSTM}
\end{figure}

% RNNs
\textbf{Recurrent cells.} The importance of each feature in the temporal attention feature vector is decided by an LSTM sub-network. Through the sequential chain structure of recurrent cells, the overall features that are generally informative for entire video sequences can be discovered. We briefly describe the inner workings of the LSTM sub-network \citep{hochreiter1997long} and how the importance of each feature for the entire video is learned, as depicted in Figure~\ref{fig:LSTM}.

% Discarding features through  "forget gate layer" and storing features with "input gate layer".
To focus on salient patterns, low intensity activations are discarded in the first operation of the recurrent cell at the \textit{forget gate layer}. A decision $f_{(t)}$ is made given the input $pool(a^{[l]})_{(t)}$ and informative features from the previous frame $h_{(t-1)}$. The features that are to be stored are decided by the product of the sigmodial ($\sigma$) \textit{input gate layer} $i_{(t)}$, and the vector of candidate values $\widetilde{C}_{(t)}$ as computed as:
\vspace{-0.5mm}
\begin{equation}
\label{eq:input_gate}
\begin{split}
    i_{(t)} =\{ \sigma(w_{i} * [h_{(t-1)},pool(a^{[l]})_{(t)}] + b_{i}) \} \quad \quad\\
    \widetilde{C}_{(t)} = \{ tanh(w_{C} * [h_{(t-1)},pool(a^{[l]})_{(t)}] + b_{C}) \} \quad
\end{split}
\end{equation}

% Cell state updates
The previous cell state $C_{(t-1)}$ is then updated based on the forget and input gates in order to ignore features that are not consistent across time and to determine the update weight. The new cell state $C_{(t)}$ is calculated as:
\vspace{-0.5mm}
\begin{equation}
\label{eq:update}
    C_{(t)} = f_{(t)} * C_{(t-1)} + i_{(t)} * \widetilde{C}_{(t)}
\end{equation}

% Cell output
The output of the recurrent cell $h_{(t)}$ is given by the current cell state $C_{(t)}$, the previous hidden state $h_{(t-1)}$ and current input $pool(a^{[l]})_{(t)}$ as:
\vspace{-0.5mm}
\begin{equation}
\label{eq:lstmout}
\begin{split}
h_{(t)} = a_{(t)} * tanh(C_{(t)}), where \qquad \qquad \\
a_{(t)} =\{ \sigma(w_{a} * [h_{(t-1)},pool(a^{[l]})_{(t)}] + b_{a}) \}
\end{split}
\end{equation}

% differences between LSTM inputs and produced activations
The hidden states are again squeezed together to re-create a coherent sequence of filtered spatio-temporal feature intensities $a^{\star[l]}$. This new attention vector considers previous cell states, thus creating a generalized vector based on the feature intensity across time.

\subsection{Temporal Gates for cyclic consistency}
\label{sec:TG}

\begin{figure}[!htb]
\centering
\includegraphics[width=0.85\linewidth]{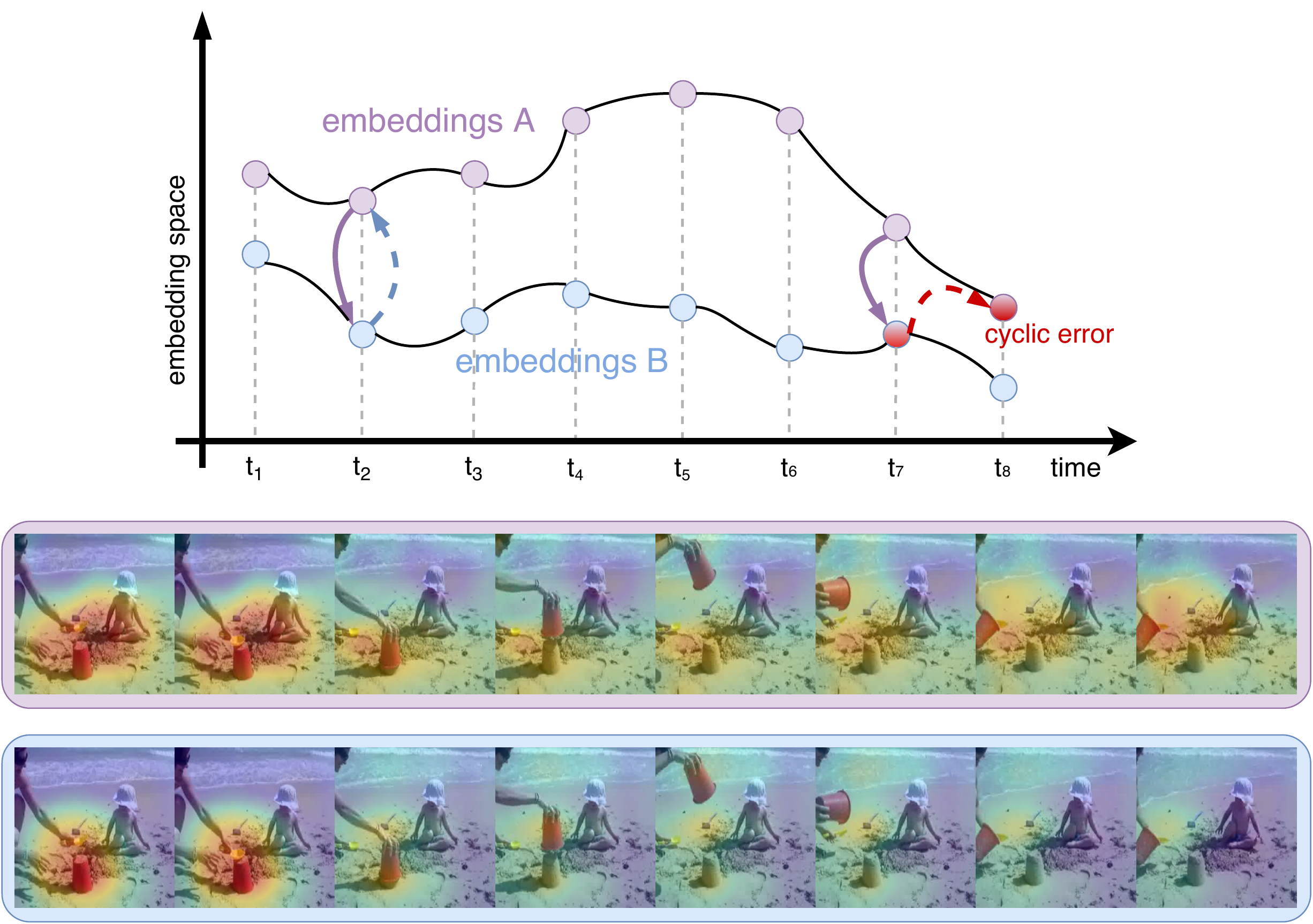}
\caption{\textbf{Temporal Cyclic Error.} Soft nearest neighbor is used to match points between two embeddings. Cycle-consistent points cycle back to original points (visualized for $t_{2}$). Otherwise, a temporal cyclic error occurs (e.g. at $t_{7}$). Corresponding salient areas below are visualized with \textit{CFP} \citep{stergiou2019class}.}
\label{fig:Cyclic_consistency}
\end{figure}

% Cyclic Consistency definition.
\textbf{Cyclic consistency.} To evaluate the similarity between two temporal volumes, cyclic consistency has been widely used \citep{dwibedi2019temporal,wang2019learning}. The technique is based on the one-to-one mapping of frames from two time sequences, schematically summarized in Figure~\ref{fig:Cyclic_consistency}. Each of the two feature spaces can be considered an \textit{embedding space}. Two embedding spaces are cycle-consistent \textbf{if and only if}, each point at time  $t$ in the embedding space \textbf{\textit{A}}, has a minimum distance point in embedding space \textbf{\textit{B}} that is also at time $t$. Equivalently, each point at time $t$ in embedding space \textbf{\textit{B}} should also have a minimum distance point in embedding space \textbf{\textit{A}} at time $t$. As shown in Figure~\ref{fig:Cyclic_consistency}, when points do not cycle back to the same temporal location, they do not exhibit cyclic consistency. In this case, a temporal cyclic error occurs.

% The importance of cyclic consistency.
By having points that can cycle back to themselves, a similarity baseline between embedding spaces can be established. Although individual features of the two spaces may be different, they should demonstrate an overall similarity as long as their alignment in terms of cyclic consistency is the same. Therefore, comparing volumes by their cyclic consistency is a suitable measure to account for (temporal) variations.

\begin{figure*}
\centering
\includegraphics[width=\textwidth]{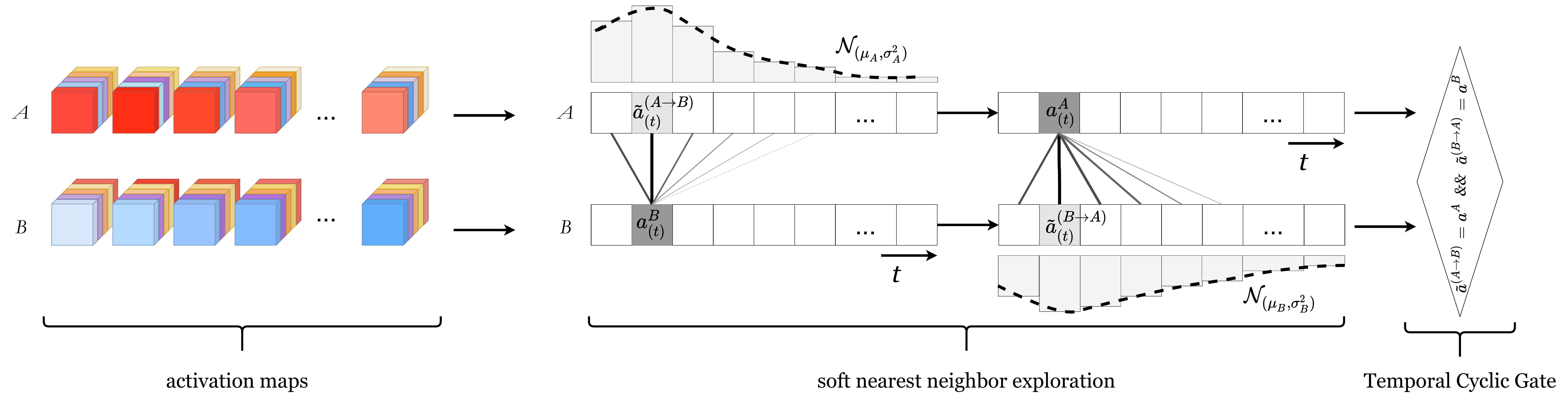}
\caption{\textbf{Temporal Gates.} Activations of each frame ($a_{(t_i)}^{B}$) in embedding space \textbf{B} are compared to the activations of every frame ($a_{(t_j)}^{A}$) in embedding space \textbf{A}. We calculate for each frame-wise activation map ($a_{(t)}^{B}$) the corresponding soft nearest neighbor ($\widetilde{a}^{A \rightarrow B}_{(t)}$) in encoding space \textbf{A}. We then equivalently obtain $\widetilde{a}_{(t)}^{B \rightarrow A}$ in encoding space \textbf{B}. The gate is open when $\widetilde{a}^{A \rightarrow B}$ and $\widetilde{a}^{B \rightarrow A}$ are exactly and sequentially equal to $a^{A}$ and $a^{B}$.}
\label{fig:TCG_pipeline}
\end{figure*}

% Cyclic soft-nearest neighbor
\textbf{Soft nearest neighbor distance.} The main challenge in creating a coherent similarity measure between two embeddings is to deal with the vast embedding spaces, as well as to discover the ``nearest'' point in an adjacent embedding. The idea of \textit{soft matches} for projected points in embeddings \citep{goldberger2005neighbourhood} is based on finding the closest point in an embedding space through the weighted sum of all possible matches and then selecting the closest actual observation.

% Main methodology
To find the soft nearest neighbor of an activation $a_{(t)}^{A}$ in embedding space \textbf{\textit{B}}, the euclidean distances between observation $a_{(t)}^{B}$ and all points in \textbf{\textit{B}} are calculated (see Figure~\ref{fig:TCG_pipeline}). Each frame is considered a separate instance for which we want to find the minimum point in the adjacent embedding space. We weight the similarity of each frame in embedding space \textbf{\textit{B}} to activation $a^{A}_{(t)}$ using a \textit{softmax} activation and by exploiting the exponential difference between activation pairs:
\vspace{-0.5mm}
\begin{equation}
    \label{eq:soft_nn}
    \widetilde{a}^{(B \rightarrow A)}_{(t)} = \sum_{i}^{T} z_{(i)}*a^{B}_{(i)}, \; where \: z_{(i)} = \frac{e^{-||a^{A}_{(t)} - a^{B}_{(i)}||^{2}}}{\sum \limits_{i}^{T} e^{-||a^{A}_{(t)} - a^{B}_{(i)}||^{2}}}
\end{equation}

% From soft nearest neighbor to embedding 2 space frames
The \textit{softmax} activation produces a normal distribution of similarities $\mathcal{N}(\mu,\,\sigma^{2})\,$, centered on the frame with the minimum distance from activation $a^{A}_{(t)}$. Based on the discovery of the nearest neighbor $\widetilde{a}^{(B \rightarrow A)}_{(t)}$, the distance to nearest frames in \textbf{\textit{B}} can then be computed. This allows the discovery of frames that are closely related to the initially considered frame $a^{A}_{(t)}$, achieved by minimizing the L2 distance from the found soft match: 
\vspace{-0.5mm}
\begin{equation}
    \label{eq:soft_nn_to_emb2}
    a^{(B \rightarrow A)}_{(t)} = \operatorname*{argmin}_{i}(||\widetilde{a}^{(B \rightarrow A)}_{(t)} - a^{B}_{(i)}||^{2}) 
\end{equation}

% Establishing cyclic consistency
We define a point as \textit{consistent} if and only if the initial temporal location $t$ matches precisely the temporal location of the computed point in embedding space \textbf{\textit{B}}, $a^{(B \rightarrow A)}_{(t)} = a^{B}_{(t)} \: \forall t \in \{1,...,T\}$. To establish a consistency check for frames in embedding space \textbf{\textit{A}}, the same procedure is repeated in reverse for every frame in embedding space \textbf{\textit{B}}, calculating the soft nearest neighbor in embedding space \textbf{\textit{A}}. The two embeddings are considered \textit{cycle-consistent} if and only if all points on both embedding spaces map back to themselves through the other embedding space: $a^{(B \rightarrow A)}_{(t)} = a^{B}_{(t)} \; and \, a^{(A \rightarrow B)}_{(t)} = a^{A}_{(t)} \: \forall t \in \{1,...,T\}$.

% Globally and locally useful information through gating
\textbf{Temporal gates.} The temporal activation vector encapsulates average feature attention over time. However, it does not enforce a precise similarity to the local spatio-temporal activations. Thus, we compute cyclic consistency between the pooled activations $pool(a^{[l]})$ and the outputted recurrent cells $a^{\star[l]}$. In this context, cyclic consistency is used as a gating mechanism to only fuse the recurrent cell hidden states with unpooled versions of the activations when the two volumes are temporally cycle-consistent. This condition ensures that only time-consistent information is added back to the network, as shown for the active states in Figure~\ref{fig:SR_pipeline}(a).

\subsection{SRTG block variants}
\label{sec:Variants}

% Purpose of different configurations
Cyclic consistency can be considered in different parts of a convolution block, and we investigate six different approaches in terms of constructing a SRTG block. In each case, the principle of global and local information fusion remains. The block configurations only differ in the relative locations of the SRTG and the LSTM input. All configurations are shown in Figure~\ref{fig:SR_pipeline}(b). Similar to networks with residual connections, we consider Simple blocks with two conv operations and Bottleneck blocks with three conv operations. Not all SRTG configurations apply to the Simple blocks.

% Start configuration
\textbf{Start.} SRTG is the very first process in the block to ensure that all operations will be based on both global and local information. The configuration can be used in both Simple and Bottleneck residual blocks.

% Top configuration
\textbf{Top.} Activations of the first convolution are used by the LSTM, with fused features being used by the final convolution. This is specific to Bottleneck blocks.

% Mid configuration
\textbf{Mid.} SRTG is added at the middle of Simple blocks and after the second convolution at Bottleneck blocks.

% End configuration
\textbf{End.} Local and global features are fused at the end of the final convolution, before the concatenation of the residual connection. This is only used in Bottleneck blocks.

% Residual configuration
\textbf{Res.} The SRTG block can also be applied to the residual connection. This transforms the residual connection to further include global spatio-temporal features and to combine those with the convolutional activations for either Simple or Bottleneck blocks.

% Final configuration
\textbf{Final.} SRTG is added at the end of the residual block, which allows for the activations to be calculated jointly with their representations across time on the entire video. This can be used in both Simple and Bottleneck blocks.

\section{Experiments and Results}
\label{sec:results}
We evaluate our approach on five action recognition benchmark datasets (Section~\ref{subsec:datasets}). We perform experiments with various ResNet backbones with various depths. Each network uses either 3D convolutions (\textit{r3d}) or (2+1)D convolutions (\textit{r(2+1)d}).

\subsection{Datasets} \label{subsec:datasets}
% Datasets used
We use five action recognition datasets for our experiments:

% HACS
\textbf{Human Action Clips and Segments} (HACS,~\cite{zhao2019hacs}) includes approximately 500K clips of 200 classes. Clips are 60-frame segments extracted from 50k unique videos.

%Kinetics
\textbf{Kinetics-700} (K-700,~\cite{carreira2019short}) is the extension of Kinetics-400/600 to 700 classes. It contains approximately 600k clips of varying duration. 

% MiT
\textbf{Moments in Time} (MiT,~\cite{monfort2018moments}) is one of the largest video datasets of human actions and activities. It includes 339 classes with approximately 800K, 3-second clips.

% UCF
\textbf{UCF-101}~\citep{soomro2012ucf101} includes 101 classes and 13k clips that vary between 2 and 14 seconds in duration.

%HMDB
\textbf{HMDB-51}~\citep{kuehne2011hmdb} contains 7K clips divided over 51 classes with at least 101 clips per class.

\subsection{Experimental settings}

% Training settings
Training was performed with a random sub-sampling of 16 frames, resized to $224 \times 224$. We adopted a multigrid training scheme \citep{wu2019multigrid} with an initial learning rate of 0.1, halved at each cycle. We used a SGD optimizer with $1e^{-6}$ weight decay and a step-wise learning rate reduction. All tested SRTG blocks incorporate stacked dual LSTMs (2 layers). For HACS, K-700 and MiT, we use the train/test splits suggested by the authors, and report on split1 for UCF-101 and HMDB-51.

\subsection{Comparison of SRTG block configurations}
\label{sec:variants_comp}
We compare the different SRTG block configurations with a 34-layer r3d and r(2+1)d. ResNets-34 contain Simple blocks with two conv layers instead of the Bottleneck blocks with three conv layers. We therefore only evaluate the Start, Mid, Res and Final configurations. Results, summarized in Table~\ref{table:configurations}, are obtained on HACS by training from scratch. All SRTG blocks perform better than their vanilla counterparts. This demonstrates the merits of our more flexible treatment of the temporal dimension. This effect appears to be stronger when the filtering is applied later. Indeed, the best performing SRTG configuration \textit{Final} achieves a top-1 accuracy improvement of 3.781\% for 3D and 4.686\% for (2+1)D convolution blocks.

% Table 1: Configuration comparison on HACS - res34
\begin{table}[!htb]
\caption{\label{tab1}Comparison of r3d-34 with SRTG configurations on HACS.}
\centering
\resizebox{0.4\textwidth}{!}{%
\begin{tabular}{cccccc}
\hline
    \multirow{2}{*}{Config} &
    \multirow{2}{*}{Gates} &
      \multicolumn{2}{c}{top-1 (\%)} &
      \multicolumn{2}{c}{top-5 (\%)} \\
    & & 3D & (2+1)D & 3D & (2+1)D \\
    \hline
    \hline
    No SRTG & \ding{55} & 74.818 & 75.703 & 92.839 & 93.571 \\
    
    Start & \ding{51} & 75.705 & 76.438 & 93.230 & 93.781 \\
    
    Mid & \ding{51} & 75.489 & 76.685 & 93.224 & 93.746 \\
    
    Res & \ding{51} & 76.703 & 77.094 & 93.307 & 93.856 \\

    Final & \ding{51} & \textbf{78.599} & \textbf{80.389} & \textbf{93.569} & \textbf{94.267} \\
\hline
\end{tabular}
}
\label{table:configurations}
\vspace{-1mm}
\end{table}

\subsection{Comparison of network architectures}
\label{sec:results::subsec:net}

% Table 2: Complete accuracy comparison on HACS, K700, MiT, UCF and HMDB
\begin{table*}[ht]
\caption{Action recognition accuracy for all five benchmark datasets. Top part of the table includes state-of-the-art models, evaluated from the trained models provided by the respective authors. Middle and bottom parts summarize the results for r3/(2+1)d with/without SRTG, respectively.}
\centering
\resizebox{\textwidth}{!}{%
\begin{tabular}{ccccccccccc}
\hline
Model & 
\multicolumn{2}{c}{HACS} &
\multicolumn{2}{c}{Kinetics-700} &
\multicolumn{2}{c}{Moments in Time} &
\multicolumn{2}{c}{UCF-101} & 
\multicolumn{2}{c}{HMDB-51}\\
&
top-1(\%) & top-5(\%) & 
top-1(\%) & top-5(\%) & 
top-1(\%) & top-5(\%) & 
top-1(\%) & top-5(\%) & 
top-1(\%) & top-5(\%)\\
\hline
\hline
I3D & 
79.948 & 94.482 &
53.015 & 69.193 & 
28.143 & 54.570 &
92.453 & 97.619 &
71.768 & 94.128\\
\hline
TSM &
N/A & N/A &
54.032 & 72.216 & 
N/A & N/A &
92.336 & 97.961 &
72.391 & 94.158\\
\hline
ir-CSN-101 &
N/A & N/A &
54.665 & 73.784 & 
N/A & N/A &
94.708 & 98.681 &
73.554 & 95.394\\
\hline
MF-Net &
N/A & N/A &
54.249 & 73.378 & 
27.286 & 48.237 &
93.863 & 98.372 &
72.654 & 94.896\\
\hline
%SF r3d-50 \textbf{(2x nets)} &
SF r3d-50 &
N/A & N/A &
56.167 & 75.569 & 
N/A & N/A &
94.619 & 98.756 &
73.291 & 95.410\\
\hline
%SF r3d-101 \textbf{(2x nets)} &
SF r3d-101 &
N/A & N/A &
\textbf{57.326} & 77.194 & 
N/A & N/A &
95.756 & 99.138 &
74.205 & 95.974\\
\hline
\hline
%r3d-18 &
%69.273 & 90.643 &
%44.452 & 62.760 & 
%N/A & N/A &
%86.035 & 94.842 &
%65.417 & 85.667\\
%\hline
r3d-34 &
74.818 & 92.839 &
46.138 & 67.108 &
24.876 & 50.104 &
89.405 & 96.883 &
69.583 & 91.833\\
\hline
r3d-50 &
78.361 & 93.763 &
49.083 & 72.541 & 
28.165 & 53.492 &
93.126 & 96.293 &
72.192 & 94.562\\
\hline
r3d-101 &
80.492 & 95.179 &
52.583 & 74.631 &
31.466 & 57.382 &
95.756 & 98.423 &
75.650 & 95.917\\
\hline
r(2+1)d-34 &
75.703 & 93.571 &
46.625 & 68.229 & 
25.614 & 52.731 &
88.956 & 96.972 &
69.205 & 90.750\\
\hline
r(2+1)d-50 &
81.340 & 94.514 &
49.927 & 73.396 & 
29.359 & 55.241 &
93.923 & 97.843 &
73.056 & 94.381\\
\hline
r(2+1)d-101 &
82.957 & 95.683 & 
52.536 & 75.177 &
N/A & N/A &
95.503 & 98.705 &
75.837 & 95.512\\
\hline
\hline
SRTG r3d-34 &
78.599 & 93.569 &
49.153 & 72.682 & 
28.549 & 52.347 &
94.799 & 98.064 &
74.319 & 94.784\\
\hline
SRTG r3d-50 &
80.362 & 95.548 &
53.522 & 74.171 & 
30.717 & 55.650 &
95.756 & 98.550 &
75.650 & 95.674\\
\hline
SRTG r3d-101 &
81.659 & 96.326 &
56.462 & 76.819 & 
\textbf{33.564} & \textbf{58.491} &
\textbf{97.325} & \textbf{99.557} &
\textbf{77.536} & \textbf{96.253}\\
\hline
SRTG r(2+1)d-34 &
80.389 & 94.267 &
49.427 & 73.233 & 
28.972 & 54.176 &
94.149 & 97.814 &
72.861 & 92.667\\
\hline
SRTG r(2+1)d-50 &
83.774 & 96.560 &
54.174 & 74.620 & 
31.603 & 56.796 &
95.675 & 98.842 &
75.297 & 95.141\\
\hline
SRTG r(2+1)d-101 &
\textbf{84.326} & \textbf{96.852} &
56.826 & \textbf{77.439} & 
N/A & N/A &
97.281 & 99.160 &
77.036 & 95.985\\
\hline
\hline
\end{tabular}%
}
\label{table:accuracies_full}
\vspace{-1mm}
\end{table*}

% Figure 6: accuracy comparisons
\begin{figure*}[!htb]%
    \centering
    \begin{subfigure}{.19\linewidth}
    	\centering\ \includegraphics[width=\linewidth]{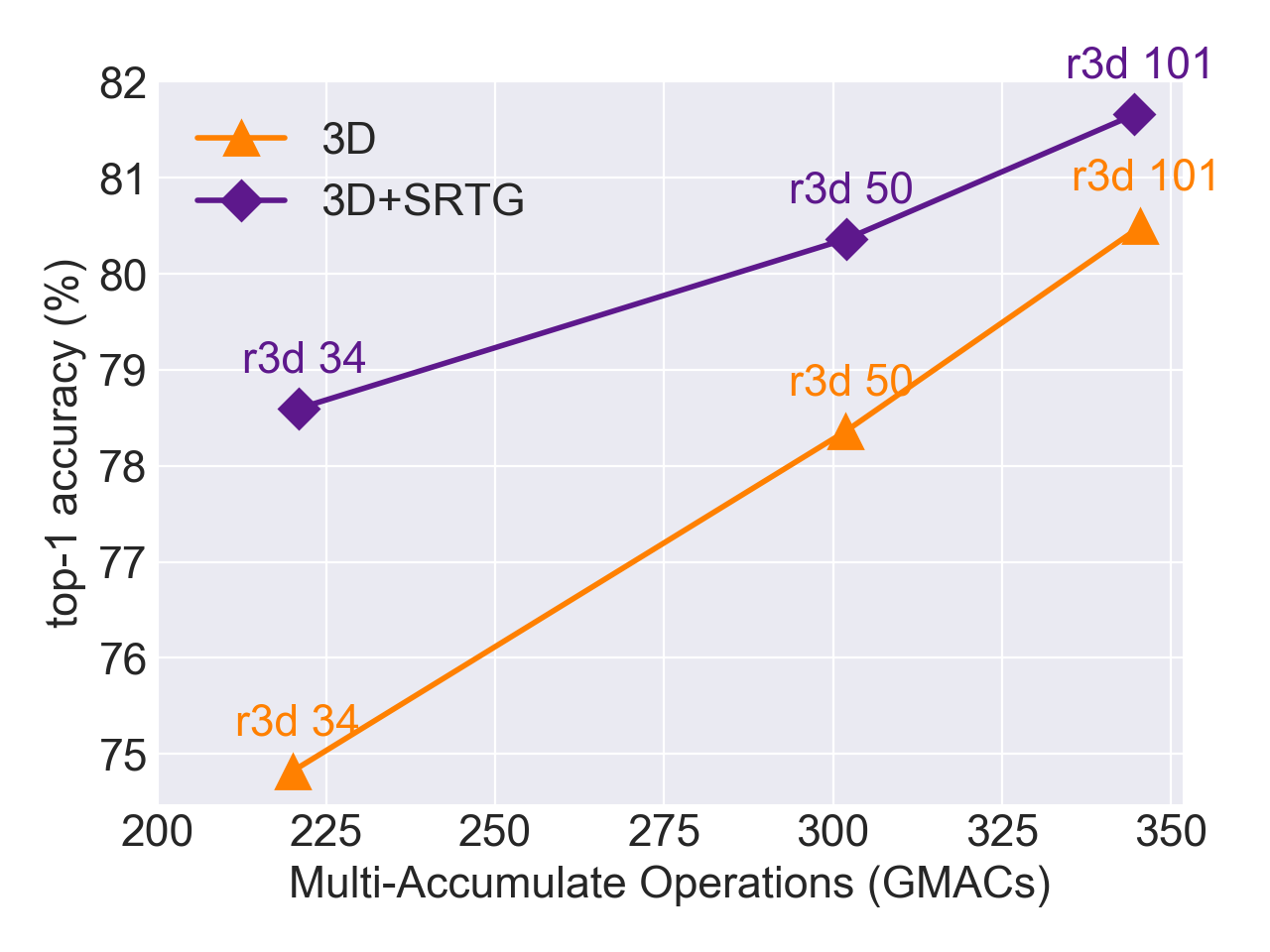}
   		\caption{HACS (r3d)}\label{fig:acc2gmacs::subfig::a}
    \end{subfigure}
    \begin{subfigure}{.19\linewidth}
    	\centering\ \includegraphics[width=\linewidth]{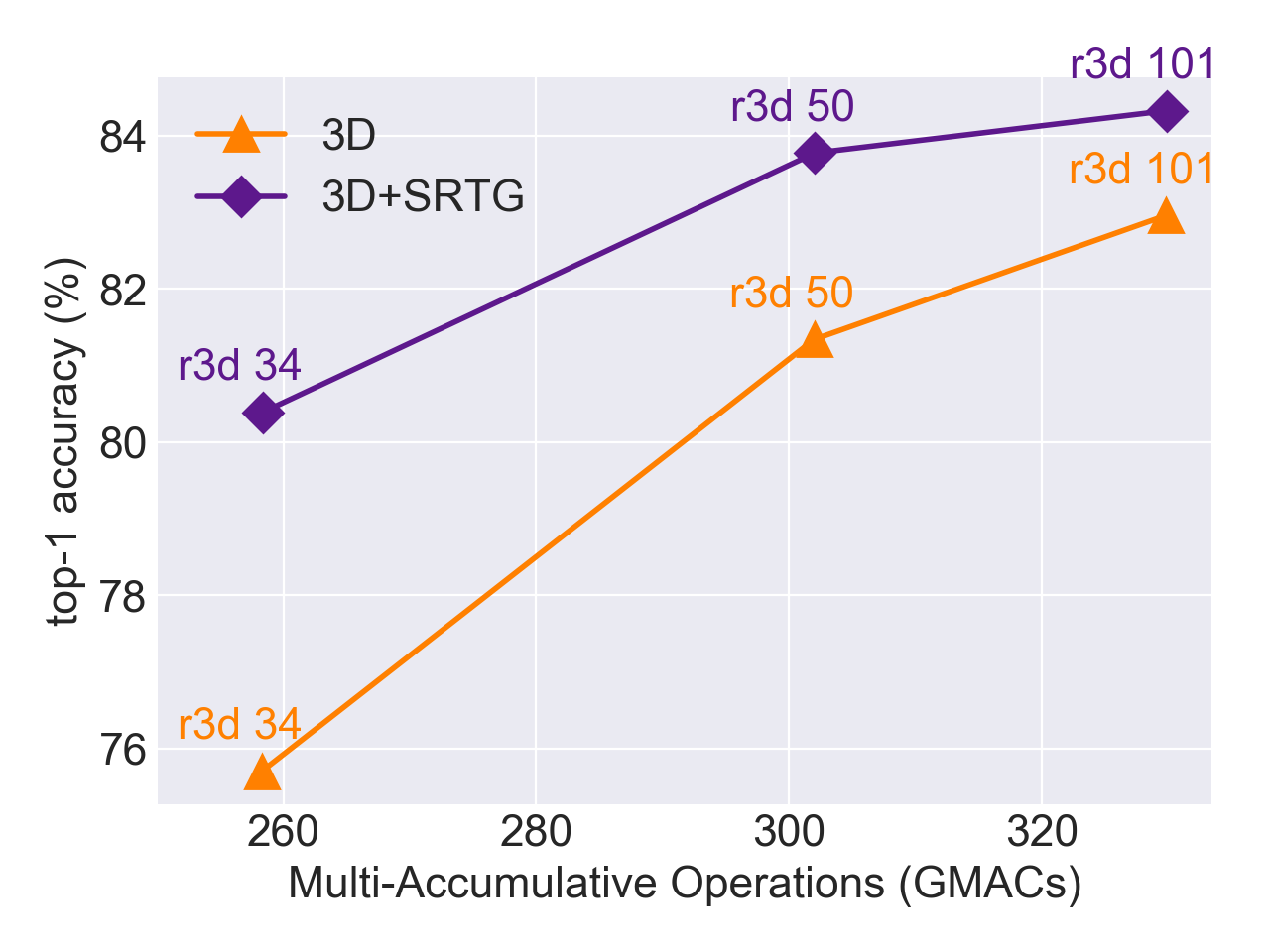}
   		\caption{HACS (r(2+1)d)}\label{fig:acc2gmacs::subfig::b}
    \end{subfigure}
    \begin{subfigure}{.19\linewidth}
    	\centering\ \includegraphics[width=\linewidth]{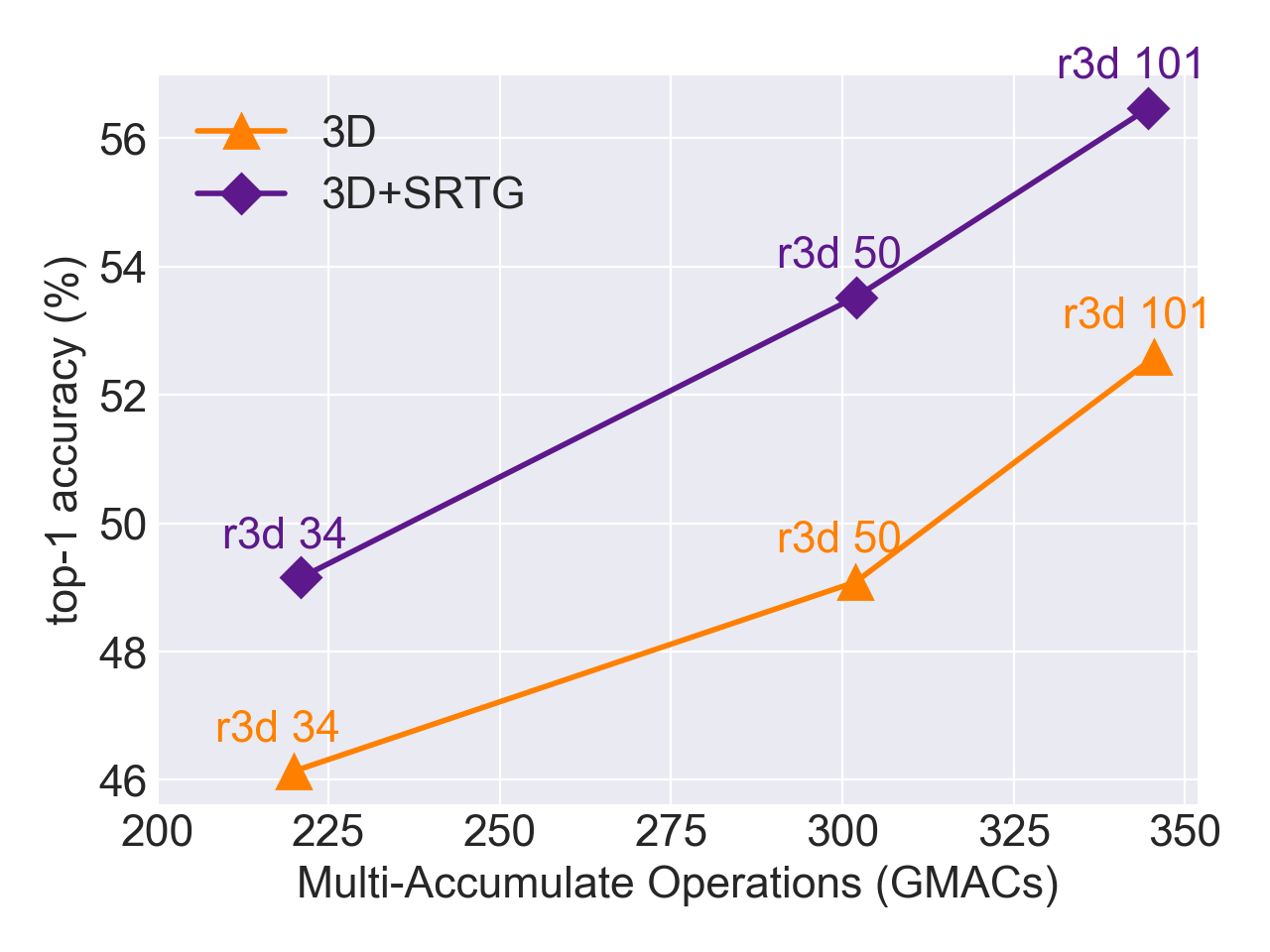}
   		\caption{K-700 (r3d)}\label{fig:acc2gmacs::subfig::c}
    \end{subfigure}
    \begin{subfigure}{.19\linewidth}
    	\centering\ \includegraphics[width=\linewidth]{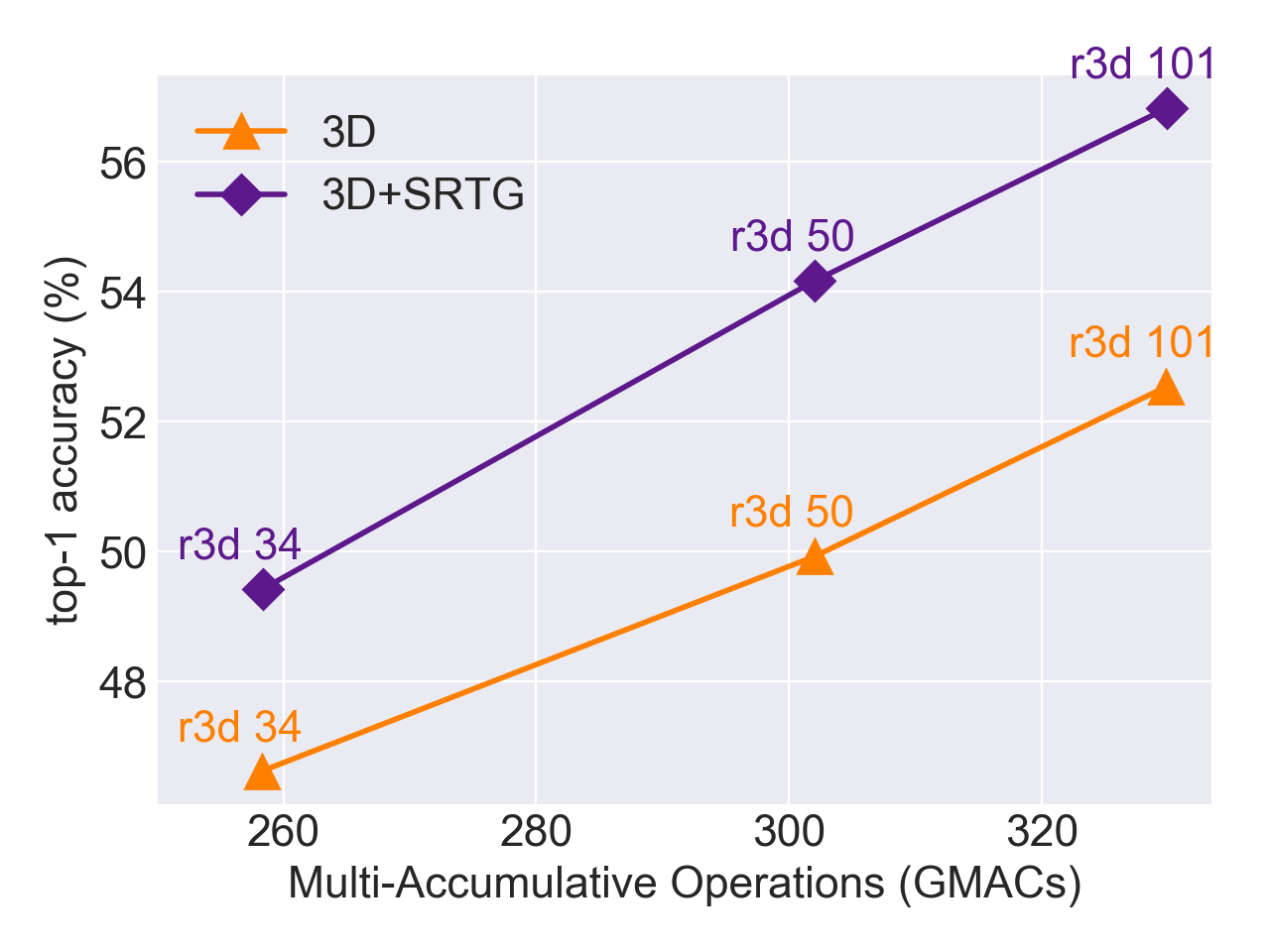}
   		\caption{K-700 (r(2+1)d)}\label{fig:acc2gmacs::subfig::d}
    \end{subfigure}
    \begin{subfigure}{.19\linewidth}
    	\centering\ \includegraphics[width=\linewidth]{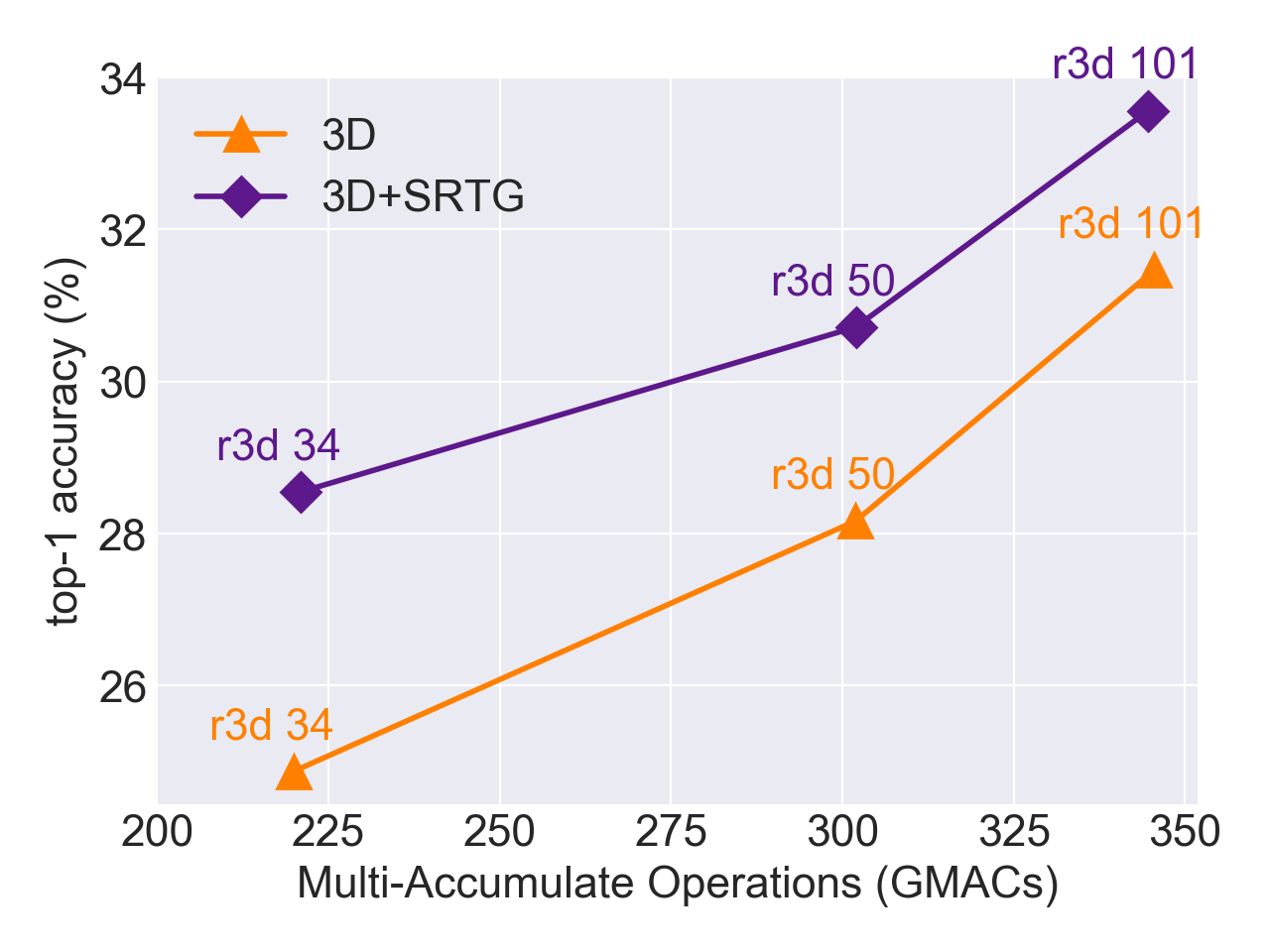}
   		\caption{MiT (r3d)}\label{fig:acc2gmacs::subfig::e}
    \end{subfigure}
    \caption{\textbf{Accuracy in relation to computation cost.} Top-1 accuracy and operations (in GMACs) of r3/r(2+1)d with/without SRTG on HACS, K-700 and MiT.}
    \label{fig:acc2gmacs}%
\vspace{-1mm}
\end{figure*}

% overall goal and table
To better understand the merits of our method, we compare a number of network architectures with and without SRTG (Final configuration). We summarize the performance on all five benchmark datasets in Table~\ref{table:accuracies_full}. The top part of the table contains the results for state-of-the-art networks including I3D \cite{carreira2017quo} which is based on an Inception-v1 network. The remaining evaluated architectures use Resnet backbones. Temporal Shift Module (TSM, \cite{lin2019tsm}) and Multi-Fiber networks (MF, \cite{chen2018multifiber}) use a r3d-50 backbone and Channel-Separated Convolutions (ir-CSN, \cite{tran2019video}) and SlowFast networks (SF, \cite{feichtenhofer2019slowfast}) are based on r3d-101 backbones. We further include an additional 50-layer SlowFast network for an additional comparison of lower-capacity models. We have used the trained networks from the respective authors' repositories. These trained models are typically pre-trained on other datasets. Missing values are due to the lack of a trained model. Any deviations from previously reported performances are due to the use of multigrid \citep{wu2019multigrid} with a base cycle batch size of 32.

The second and third parts of Table~\ref{table:accuracies_full} summarize the performances of ResNets with various depths and 3D or (2+1)D convolutions, with and without SRTG, respectively. Models for HACS are trained from scratch. The weights of models for K-700 and MiT are initialized based on those from the pre-trained HACS model. For UCF-101 and HMDB-51, we fine-tune the HACS pre-trained models. Missing values are due to time constraints. We will add these in the final version of the paper.
% TODO RONALD: replace last two sentences upon acceptance

% general observations
For the state-of-the-art architectures, the use of larger and deeper models provides accuracy improvements. This is in line with the general trend for action recognition using CNNs with architectures that are either deeper or include higher complexity. Models implemented with (2+1)D convolution blocks perform somewhat better than their counterparts with 3D convolutions. These differences are modest and not consistent across datasets, however.

% SRTG
%As shown in Table~\ref{table:accuracies_full}, adding SRTG blocks to any architecture consistently improves performance. Table~\ref{table:blockcomparisons} shows pairwise comparisons of the performance on the three largest benchmark datasets for networks with and without SRTG. On average, these improvement are approximately \textcolor{red}{4.13\%} for Kinetics-700, 2.6\% for MiT and 2.57\% for HACS, and independent of the convolution block type. For smaller networks, the performance gains are somewhat higher even with average improvements of 4.12\% for r3d-34, 3.94\% for r(2+1)d-34, 3.016\% for r3d-50 and 2.59\% for r(2+1)d-50 over all five datasets. Clearly, the use of time-consistent features obtained through our method improves a generalization ability of 3D-CNNs.
As shown in Table~\ref{table:accuracies_full}, adding SRTG blocks to any architecture consistently improves performance. Table~\ref{table:blockcomparisons} shows pairwise comparisons of the performance on the three largest benchmark datasets for networks with and without SRTG. When using SRTG blocks, the improvements are in the range of 1.2--4.7\% for HACS, 2.8--4.4\% for K-700 and 2.1--3.7\% for MiT. For smaller networks, we observe larger gains. The use of time-consistent features obtained through our method appears to improve the generalization ability of 3D-CNNs.

% outperform
The r3d and r(2+1)d networks with SRTG perform on-par with the current state-of-the-art architectures. The r3d-101 outperforms current state-of-the-art in HACS, MiT, UCF-101 and HMDB-51. For MiT, our top-1 accuracy of 33.564\%, which largely surpasses other tested architectures. The (2+1)D variant further outperforms current architectures on HACS with 84.326\% top-1 accuracy. We also note a performance on Kinectics-700 that is comparable to the best performing SlowFast r3d-101 model. While the SlowFast network achieves better top-1 accuracy, a r(2+1)d-101 network with SRTG blocks has higher top-5 accuracy. This similar performance is remarkable given the relatively low complexity of the SRTG r3d-101 and r(2+1)d-101 models. SlowFast is built on a dual-network configuration with two sub-parts responsible for long-term and short-term movements. The SlowFast network therefore includes a significantly larger number of operations than a r3d-101 or r(2+1)d network with plug-in SRTG blocks. We analyze the computation cost of the SRTG block in Section~\ref{sec:results::subsec:conv}.

% datasets
Finally, we observe that the performance gain with SRTG is substantial for the two smaller datasets, UCF-101 and HMDB-51. Especially for UCF-101, the action recognition accuracy is very saturated. Still, the already competitive performance of the ResNet-101 models on UCF-101 increases with 1.569\% and 1.778\% for the 3D and (2+1)D convolution variants, respectively. This further demonstrates that SRTG can improve the selection of features that contain less noise and generalize better, even when there is fewer training data available.

\subsection{Analysis of computational overhead}
\label{sec:results::subsec:conv}
% Accuracy to GMACs
The SRTG block can be added to a large range of 3D-CNN architectures. It leverages the small computational costs of LSTMs compared to 3D convolutions. That enables us to increase the number of parameters without a significant increase in the number of GFLOPs. This also corresponds to the small additional memory usage compared to baseline models on both forward and backward passes. We present the number of multi-accumulative operations (MACs)\footnotemark~ used for the r3/(2+1)d architectures with and without SRTG in Figure~\ref{fig:acc2gmacs}, with respect to the corresponding accuracies. The additional computation overhead, for models that include the proposed block, is approximately 0.15\% of the total number of operations in the vanilla networks. This constitutes a negligible increase, compared to the performance gains, making SRTG a lightweight block that can be easily used on top of networks.

\footnotetext{Multi-accumulative operations \citep{ludgate1982proposed} are based on the product of two numbers increased by an accumulator. They relate to the accumulated sum of convolutions between the dot product of the weights and input region.}

% Figure 7
%\begin{figure*}
%\centering
%\includegraphics[width=\textwidth]{kinetics_class_accuracies.pdf}
%\caption{\textbf{Individual class accuracies on Kinetics-700.} The r3d-50 model was used as backbone for comparing accuracies for architectures without (denoted with \textcolor{Orange}{orange}) and with (denoted with \textcolor{Plum}{purple}) the proposed \textit{Squeezed and Recursion Temporal Gates}.}
%\label{fig:kinetics_acc}
%\end{figure*}

% Table 3: 3D block comparison (3D and 2+1D) - with/without SRTG
\begin{table}[ht]
\vspace{-1mm}
\caption{Pairwise comparisons of r3d and r(2+1)d networks with and without SRTG on HACS, K-700 and MiT.}
\vspace{-1.5mm}
\centering
\resizebox{\linewidth}{!}{%
\begin{tabular}{ccccccc}
\hline
Dataset & 
\multicolumn{2}{c}{r3d-50} &
\multicolumn{2}{c}{r(2+1)d-50} &
\multicolumn{2}{c}{r3d-101} \\
& \textcolor{Orange}{\textbf{None}} & \textcolor{Plum}{\textbf{SRTG}} & \textcolor{Orange}{\textbf{None}} & \textcolor{Plum}{\textbf{SRTG}} & \textcolor{Orange}{\textbf{None}} & \textcolor{Plum}{\textbf{SRTG}} \\
\hline
HACS &  \textcolor{Orange}{78.361} & \textcolor{Plum}{80.362 \textbf{(+2.0)}} &
        \textcolor{Orange}{81.340} & \textcolor{Plum}{83.474 \textbf{(+2.1)}} & 
        \textcolor{Orange}{80.492} & \textcolor{Plum}{81.659 \textbf{(+1.1)}}\\
K-700 & \textcolor{Orange}{49.083} & \textcolor{Plum}{53.522 \textbf{(+4.4)}} &
        \textcolor{Orange}{49.927} & \textcolor{Plum}{54.174  \textbf{(+4.2)}} &
        \textcolor{Orange}{52.583} & \textcolor{Plum}{56.462 \textbf{(+3.8)}} \\
MiT &   \textcolor{Orange}{28.165} & \textcolor{Plum}{30.717 \textbf{(+2.5)}} &
        \textcolor{Orange}{29.359} & \textcolor{Plum}{31.603 \textbf{(+3.3)}} & 
        \textcolor{Orange}{31.466} & \textcolor{Plum}{33.564 \textbf{(+2.0)}} \\
        
\end{tabular}
}
\label{table:blockcomparisons}
\vspace{-2mm}
\end{table}

\subsection{Evaluating feature transferability}
\label{sec:results::subsec:tl}
A common practice to train CNNs is to use transfer learning on a pre-trained network. To evaluate the performance of the SRTG block after transfer learning, we pre-train on several datasets and fine-tune on smaller datasets UCF-101 and HMDB-51. Through this, we can further eliminate biases relating to the pre-training datsets and compare the accuracies achieved with respect to the SRTG blocks.

As shown in Table~\ref{tab:TL}, the accuracy rates remain fairly consistent for the pre-training datasets. This consistency is due to the large sizes of these datasets, as well as the overall robustness of the proposed method. The average offset between each of the pre-trained models is 0.71\% for UCF-101 and 0.47\% for HMDB-51. These are only minor changes in accuracy, which further demonstrates that the improvements observed are due to the inclusion of SRTG blocks in the network.

% Table 4: Transfer learning (With/Without)
\begin{table}[ht]
\caption{Results on UCF-101 and HMDB-51 based on transfer learning.}
\centering
\resizebox{\linewidth}{!}{%
\begin{tabular}{ccccccccccc}
\hline
Model & Pre-training & GFLOPs & UCF-101 top-1 (\%) & HMDB-51 top-1 (\%)\\
\hline
\multirow{3}{*}{SRTG r3d-34} & HACS & \multirow{3}{*}{110.48} & 94.799 & 74.319 \\
& HACS+K-700 & & 95.842 & 74.183 \\
& HACS+MiT & & 95.166 & 74.235 \\
\hline
\multirow{3}{*}{SRTG r(2+1)d-34} & HACS & \multirow{3}{*}{110.8} & 94.149 & 72.861 \\
& HACS+K-700 &  & 94.569 & 73.217 \\
& HACS+MiT & & 95.648 & 74.473 \\
\hline
\multirow{3}{*}{SRTG r3d-50} & HACS & \multirow{3}{*}{150.98} & 95.756 & 75.650 \\
& HACS+K-700 & & 96.853 & 75.972 \\
& HACS+MiT & & 96.533 & 76.014 \\
\hline
\multirow{3}{*}{SRTG r(2+1)d-50} & HACS & \multirow{3}{*}{151.6} & 95.675 & 75.297 \\
& HACS+K-700 & & 95.993 & 75.743 \\
& HACS+MiT & & 96.278 & 75.988 \\
\hline
%\multirow{3}{*}{SRTG r3d-101} & HACS & \multirow{3}{*}{171.02} & 97.325 & 77.536 \\
%& HACS+K-700 & & 97.404 & 78.026 \\
%& HACS+MiT & & 97.568 & 78.419 \\
%\hline
\end{tabular}
}
\vspace{-1mm}
\label{tab:TL}
\end{table}

\section{Conclusions}
\label{sec:conclusions}

We have introduced a novel Squeeze and Recursion Temporal Gates (SRTG) block that can be added to a large range of CNN architectures to create time-consistent features. The SRTG block uses an LSTM to capture multi-frame feature dynamics, and a temporal gate to evaluate the cyclic consistency between the discovered dynamics and the modeled features. SRTG blocks add a negligible computational overhead (0.03--0.4 GFLOPs), which makes both forward and backward passes efficient. Adding our proposed SRTG blocks in ResNet backbones with 3D or (2+1)D convolutions consistently leads to performance gains.
% TODO: option 1
We obtain results that are on par with, and in most cases outperform, the current state-of-the-art on action recognition datasets including Kinetics-700 and Moments in Time. For HACS, we obtain a state-of-the-art-performance of 84.3\%. Our combined experiments demonstrate the generalization ability of the discovered time-consistent features.
% TODO: option 2
%We obtain results that are on par with, and in most cases outperform, the current state-of-the-art on action recognition datasets including HACS, Kinetics-700 and Moments in Time. We even show 3.05\% improvement on the almost saturated UCF-101 dataset, which further demonstrates the generalization ability of the discovered time-consistent features.
% Improvement HACS: TODO - TODO (3.05%)

\section{Acknowledgments}
\label{sec:acknowledgments}
This publication is supported by the Netherlands Organization for Scientific Research (NWO) with a TOP-C2 grant for “Automatic recognition of bodily interactions” (ARBITER).

\bibliographystyle{model2-names}
\bibliography{refs}

\begin{thebibliography}{33}
\expandafter\ifx\csname natexlab\endcsname\relax\def\natexlab#1{#1}\fi
\providecommand{\url}[1]{\texttt{#1}}
\providecommand{\href}[2]{#2}
\providecommand{\path}[1]{#1}
\providecommand{\DOIprefix}{doi:}
\providecommand{\ArXivprefix}{arXiv:}
\providecommand{\URLprefix}{URL: }
\providecommand{\Pubmedprefix}{pmid:}
\providecommand{\doi}[1]{\href{http://dx.doi.org/#1}{\path{#1}}}
\providecommand{\Pubmed}[1]{\href{pmid:#1}{\path{#1}}}
\providecommand{\bibinfo}[2]{#2}
\ifx\xfnm\relax \def\xfnm[#1]{\unskip,\space#1}\fi
%Type = Article
\bibitem[{Carreira et~al.(2019)Carreira, Noland, Hillier and
  Zisserman}]{carreira2019short}
\bibinfo{author}{Carreira, J.}, \bibinfo{author}{Noland, E.},
  \bibinfo{author}{Hillier, C.}, \bibinfo{author}{Zisserman, A.},
  \bibinfo{year}{2019}.
\newblock \bibinfo{title}{A short note on the {Kinetics-700} human action
  dataset}.
\newblock \bibinfo{journal}{arXiv preprint arXiv:1907.06987} .
%Type = Inproceedings
\bibitem[{Carreira and Zisserman(2017)}]{carreira2017quo}
\bibinfo{author}{Carreira, J.}, \bibinfo{author}{Zisserman, A.},
  \bibinfo{year}{2017}.
\newblock \bibinfo{title}{Quo vadis, action recognition? {A} new model and the
  {K}inetics dataset}, in: \bibinfo{booktitle}{Computer Vision and Pattern
  Recognition (CVPR)}, \bibinfo{organization}{IEEE}. pp.
  \bibinfo{pages}{4724--4733}.
%Type = Inproceedings
\bibitem[{Chen et~al.(2018)Chen, Kalantidis, Li, Yan and
  Feng}]{chen2018multifiber}
\bibinfo{author}{Chen, Y.}, \bibinfo{author}{Kalantidis, Y.},
  \bibinfo{author}{Li, J.}, \bibinfo{author}{Yan, S.}, \bibinfo{author}{Feng,
  J.}, \bibinfo{year}{2018}.
\newblock \bibinfo{title}{Multi-fiber networks for video recognition}, in:
  \bibinfo{booktitle}{European Conference on Computer Vision (ECCV)}, pp.
  \bibinfo{pages}{352--367}.
%Type = Inproceedings
\bibitem[{Dwibedi et~al.(2019)Dwibedi, Aytar, Tompson, Sermanet and
  Zisserman}]{dwibedi2019temporal}
\bibinfo{author}{Dwibedi, D.}, \bibinfo{author}{Aytar, Y.},
  \bibinfo{author}{Tompson, J.}, \bibinfo{author}{Sermanet, P.},
  \bibinfo{author}{Zisserman, A.}, \bibinfo{year}{2019}.
\newblock \bibinfo{title}{Temporal cycle-consistency learning}, in:
  \bibinfo{booktitle}{Conference on Computer Vision and Pattern Recognition
  (CVPR)}, \bibinfo{organization}{IEEE}. pp. \bibinfo{pages}{1801--1810}.
%Type = Article
\bibitem[{Feichtenhofer(2020)}]{feichtenhofer2020x3d}
\bibinfo{author}{Feichtenhofer, C.}, \bibinfo{year}{2020}.
\newblock \bibinfo{title}{{X3D}: Expanding architectures for efficient video
  recognition}.
\newblock \bibinfo{journal}{arXiv preprint arxiv:2004.04730} .
%Type = Inproceedings
\bibitem[{Feichtenhofer et~al.(2019)Feichtenhofer, Fan, Malik and
  He}]{feichtenhofer2019slowfast}
\bibinfo{author}{Feichtenhofer, C.}, \bibinfo{author}{Fan, H.},
  \bibinfo{author}{Malik, J.}, \bibinfo{author}{He, K.}, \bibinfo{year}{2019}.
\newblock \bibinfo{title}{{SlowFast} networks for video recognition}, in:
  \bibinfo{booktitle}{International Conference on Computer Vision (ICCV)},
  \bibinfo{organization}{IEEE}. pp. \bibinfo{pages}{6202--6211}.
%Type = Inproceedings
\bibitem[{Goldberger et~al.(2005)Goldberger, Hinton, Roweis and
  Salakhutdinov}]{goldberger2005neighbourhood}
\bibinfo{author}{Goldberger, J.}, \bibinfo{author}{Hinton, G.E.},
  \bibinfo{author}{Roweis, S.T.}, \bibinfo{author}{Salakhutdinov, R.R.},
  \bibinfo{year}{2005}.
\newblock \bibinfo{title}{Neighbourhood components analysis}, in:
  \bibinfo{booktitle}{Advances in neural information processing systems
  (NIPS)}, pp. \bibinfo{pages}{513--520}.
%Type = Inproceedings
\bibitem[{Hara et~al.(2018)Hara, Kataoka and Satoh}]{hara2018can}
\bibinfo{author}{Hara, K.}, \bibinfo{author}{Kataoka, H.},
  \bibinfo{author}{Satoh, Y.}, \bibinfo{year}{2018}.
\newblock \bibinfo{title}{Can spatiotemporal {3D CNNs} retrace the history of
  {2D CNNs} and {I}mage{N}et?}, in: \bibinfo{booktitle}{Computer Vision and
  Pattern Recognition (CVPR)}, \bibinfo{organization}{IEEE}. pp.
  \bibinfo{pages}{18--22}.
%Type = Article
\bibitem[{Herath et~al.(2017)Herath, Harandi and Porikli}]{herath2017going}
\bibinfo{author}{Herath, S.}, \bibinfo{author}{Harandi, M.},
  \bibinfo{author}{Porikli, F.}, \bibinfo{year}{2017}.
\newblock \bibinfo{title}{Going deeper into action recognition: A survey}.
\newblock \bibinfo{journal}{Image and vision computing} \bibinfo{volume}{60},
  \bibinfo{pages}{4--21}.
%Type = Article
\bibitem[{Hochreiter and Schmidhuber(1997)}]{hochreiter1997long}
\bibinfo{author}{Hochreiter, S.}, \bibinfo{author}{Schmidhuber, J.},
  \bibinfo{year}{1997}.
\newblock \bibinfo{title}{Long short-term memory}.
\newblock \bibinfo{journal}{Neural computation} \bibinfo{volume}{9},
  \bibinfo{pages}{1735--1780}.
%Type = Inproceedings
\bibitem[{Hu et~al.(2018a)Hu, Shen, Albanie, Sun and Vedaldi}]{hu2018gather}
\bibinfo{author}{Hu, J.}, \bibinfo{author}{Shen, L.}, \bibinfo{author}{Albanie,
  S.}, \bibinfo{author}{Sun, G.}, \bibinfo{author}{Vedaldi, A.},
  \bibinfo{year}{2018}a.
\newblock \bibinfo{title}{Gather-excite: Exploiting feature context in
  convolutional neural networks}, in: \bibinfo{booktitle}{Advances in Neural
  Information Processing Systems (NIPS)}, pp. \bibinfo{pages}{9401--9411}.
%Type = Inproceedings
\bibitem[{Hu et~al.(2018b)Hu, Shen and Sun}]{hu2018squeeze}
\bibinfo{author}{Hu, J.}, \bibinfo{author}{Shen, L.}, \bibinfo{author}{Sun,
  G.}, \bibinfo{year}{2018}b.
\newblock \bibinfo{title}{Squeeze-and-excitation networks}, in:
  \bibinfo{booktitle}{Conference on Computer Vision and Pattern Recognition
  (CVPR)}, \bibinfo{organization}{IEEE}. pp. \bibinfo{pages}{7132--7141}.
%Type = Article
\bibitem[{Ji et~al.(2013)Ji, Xu, Yang and Yu}]{ji20133d}
\bibinfo{author}{Ji, S.}, \bibinfo{author}{Xu, W.}, \bibinfo{author}{Yang, M.},
  \bibinfo{author}{Yu, K.}, \bibinfo{year}{2013}.
\newblock \bibinfo{title}{{3D} convolutional neural networks for human action
  recognition}.
\newblock \bibinfo{journal}{Transactions on Pattern Analysis and Machine
  Intelligence} \bibinfo{volume}{35}, \bibinfo{pages}{221--231}.
%Type = Inproceedings
\bibitem[{Kuehne et~al.(2011)Kuehne, Jhuang, Garrote, Poggio and
  Serre}]{kuehne2011hmdb}
\bibinfo{author}{Kuehne, H.}, \bibinfo{author}{Jhuang, H.},
  \bibinfo{author}{Garrote, E.}, \bibinfo{author}{Poggio, T.},
  \bibinfo{author}{Serre, T.}, \bibinfo{year}{2011}.
\newblock \bibinfo{title}{{HMDB}: {A} large video database for human motion
  recognition}, in: \bibinfo{booktitle}{International Conference on Computer
  Vision (ICCV)}, \bibinfo{organization}{IEEE}. pp.
  \bibinfo{pages}{2556--2563}.
%Type = Inproceedings
\bibitem[{Lin et~al.(2019)Lin, Gan and Han}]{lin2019tsm}
\bibinfo{author}{Lin, J.}, \bibinfo{author}{Gan, C.}, \bibinfo{author}{Han,
  S.}, \bibinfo{year}{2019}.
\newblock \bibinfo{title}{{TSM}: Temporal shift module for efficient video
  understanding}, in: \bibinfo{booktitle}{International Conference on Computer
  Vision (ICCV)}, \bibinfo{organization}{IEEE}. pp.
  \bibinfo{pages}{7083--7093}.
%Type = Inproceedings
\bibitem[{Long et~al.(2018)Long, Gan, De~Melo, Wu, Liu and
  Wen}]{long2018attention}
\bibinfo{author}{Long, X.}, \bibinfo{author}{Gan, C.},
  \bibinfo{author}{De~Melo, G.}, \bibinfo{author}{Wu, J.},
  \bibinfo{author}{Liu, X.}, \bibinfo{author}{Wen, S.}, \bibinfo{year}{2018}.
\newblock \bibinfo{title}{Attention clusters: Purely attention based local
  feature integration for video classification}, in:
  \bibinfo{booktitle}{Conference on Computer Vision and Pattern Recognition
  (CVPR)}, \bibinfo{organization}{IEEE}. pp. \bibinfo{pages}{7834--7843}.
%Type = Incollection
\bibitem[{Ludgate(1982)}]{ludgate1982proposed}
\bibinfo{author}{Ludgate, P.E.}, \bibinfo{year}{1982}.
\newblock \bibinfo{title}{On a proposed analytical machine}, in:
  \bibinfo{booktitle}{The Origins of Digital Computers}.
  \bibinfo{publisher}{Springer}, pp. \bibinfo{pages}{73--87}.
%Type = Article
\bibitem[{Monfort et~al.(2019)Monfort, Andonian, Zhou, Ramakrishnan, Bargal,
  Yan, Brown, Fan, Gutfreund, Vondrick et~al.}]{monfort2018moments}
\bibinfo{author}{Monfort, M.}, \bibinfo{author}{Andonian, A.},
  \bibinfo{author}{Zhou, B.}, \bibinfo{author}{Ramakrishnan, K.},
  \bibinfo{author}{Bargal, S.A.}, \bibinfo{author}{Yan, T.},
  \bibinfo{author}{Brown, L.}, \bibinfo{author}{Fan, Q.},
  \bibinfo{author}{Gutfreund, D.}, \bibinfo{author}{Vondrick, C.}, et~al.,
  \bibinfo{year}{2019}.
\newblock \bibinfo{title}{Moments in time dataset: {O}ne million videos for
  event understanding}.
\newblock \bibinfo{journal}{IEEE Transactions on Pattern Analysis and Machine
  Intelligence} \bibinfo{volume}{42}, \bibinfo{pages}{502--508}.
%Type = Inproceedings
\bibitem[{Qiu et~al.(2017)Qiu, Yao and Mei}]{qiu2017learning}
\bibinfo{author}{Qiu, Z.}, \bibinfo{author}{Yao, T.}, \bibinfo{author}{Mei,
  T.}, \bibinfo{year}{2017}.
\newblock \bibinfo{title}{Learning spatio-temporal representation with
  pseudo-{3D} residual networks}, in: \bibinfo{booktitle}{International
  Conference on Computer Vision (ICCV)}, \bibinfo{organization}{IEEE}. pp.
  \bibinfo{pages}{5534--5542}.
%Type = Inproceedings
\bibitem[{Qiu et~al.(2019)Qiu, Yao, Ngo, Tian and Mei}]{qiu2019learning}
\bibinfo{author}{Qiu, Z.}, \bibinfo{author}{Yao, T.}, \bibinfo{author}{Ngo,
  C.W.}, \bibinfo{author}{Tian, X.}, \bibinfo{author}{Mei, T.},
  \bibinfo{year}{2019}.
\newblock \bibinfo{title}{Learning spatio-temporal representation with local
  and global diffusion}, in: \bibinfo{booktitle}{Conference on Computer Vision
  and Pattern Recognition (CVPR)}, \bibinfo{organization}{IEEE}. pp.
  \bibinfo{pages}{12056--12065}.
%Type = Inproceedings
\bibitem[{Simonyan and Zisserman(2014)}]{simonyan2014two}
\bibinfo{author}{Simonyan, K.}, \bibinfo{author}{Zisserman, A.},
  \bibinfo{year}{2014}.
\newblock \bibinfo{title}{Two-stream convolutional networks for action
  recognition in videos}, in: \bibinfo{booktitle}{Advances in Neural
  Information Processing Systems (NIPS)}, pp. \bibinfo{pages}{568--576}.
%Type = Article
\bibitem[{Soomro et~al.(2012)Soomro, Zamir and Shah}]{soomro2012ucf101}
\bibinfo{author}{Soomro, K.}, \bibinfo{author}{Zamir, A.R.},
  \bibinfo{author}{Shah, M.}, \bibinfo{year}{2012}.
\newblock \bibinfo{title}{{UCF101}: A dataset of 101 human actions classes from
  videos in the wild}.
\newblock \bibinfo{journal}{arXiv preprint arXiv:1212.0402} .
%Type = Inproceedings
\bibitem[{Stergiou et~al.(2019)Stergiou, Kapidis, Kalliatakis, Chrysoulas,
  Poppe and Veltkamp}]{stergiou2019class}
\bibinfo{author}{Stergiou, A.}, \bibinfo{author}{Kapidis, G.},
  \bibinfo{author}{Kalliatakis, G.}, \bibinfo{author}{Chrysoulas, C.},
  \bibinfo{author}{Poppe, R.}, \bibinfo{author}{Veltkamp, R.},
  \bibinfo{year}{2019}.
\newblock \bibinfo{title}{Class feature pyramids for video explanation}, in:
  \bibinfo{booktitle}{International Conference on Computer Vision Workshop
  (ICCVW)}, \bibinfo{organization}{IEEE}. pp. \bibinfo{pages}{4255--4264}.
%Type = Article
\bibitem[{Stergiou and Poppe(2019a)}]{stergiou2019analyzing}
\bibinfo{author}{Stergiou, A.}, \bibinfo{author}{Poppe, R.},
  \bibinfo{year}{2019}a.
\newblock \bibinfo{title}{Analyzing human-human interactions: A survey}.
\newblock \bibinfo{journal}{Computer Vision and Image Understanding}
  \bibinfo{volume}{188}, \bibinfo{pages}{102799}.
%Type = Inproceedings
\bibitem[{Stergiou and Poppe(2019b)}]{stergiou2019FAST}
\bibinfo{author}{Stergiou, A.}, \bibinfo{author}{Poppe, R.},
  \bibinfo{year}{2019}b.
\newblock \bibinfo{title}{Spatio-temporal {FAST 3D} convolutions for human
  action recognition}, in: \bibinfo{booktitle}{International Conference on
  Machine Learning Applications (ICMLA)}, \bibinfo{organization}{IEEE}. pp.
  \bibinfo{pages}{1830--1834}.
%Type = Inproceedings
\bibitem[{Tran et~al.(2019)Tran, Wang, Torresani and Feiszli}]{tran2019video}
\bibinfo{author}{Tran, D.}, \bibinfo{author}{Wang, H.},
  \bibinfo{author}{Torresani, L.}, \bibinfo{author}{Feiszli, M.},
  \bibinfo{year}{2019}.
\newblock \bibinfo{title}{Video classification with channel-separated
  convolutional networks}, in: \bibinfo{booktitle}{International Conference on
  Computer Vision (ICCV)}, \bibinfo{organization}{IEEE}. pp.
  \bibinfo{pages}{5552--5561}.
%Type = Inproceedings
\bibitem[{Tran et~al.(2018)Tran, Wang, Torresani, Ray, LeCun and
  Paluri}]{tran2018closer}
\bibinfo{author}{Tran, D.}, \bibinfo{author}{Wang, H.},
  \bibinfo{author}{Torresani, L.}, \bibinfo{author}{Ray, J.},
  \bibinfo{author}{LeCun, Y.}, \bibinfo{author}{Paluri, M.},
  \bibinfo{year}{2018}.
\newblock \bibinfo{title}{A closer look at spatiotemporal convolutions for
  action recognition}, in: \bibinfo{booktitle}{Conference on Computer Vision
  and Pattern Recognition (CVPR)}, \bibinfo{organization}{IEEE}. pp.
  \bibinfo{pages}{6450--6459}.
%Type = Inproceedings
\bibitem[{Wang et~al.(2018a)Wang, Li, Li and Van~Gool}]{wang2018appearance}
\bibinfo{author}{Wang, L.}, \bibinfo{author}{Li, W.}, \bibinfo{author}{Li, W.},
  \bibinfo{author}{Van~Gool, L.}, \bibinfo{year}{2018}a.
\newblock \bibinfo{title}{Appearance-and-relation networks for video
  classification}, in: \bibinfo{booktitle}{Conference on Computer Vision and
  Pattern Recognition (CVPR)}, \bibinfo{organization}{IEEE}. pp.
  \bibinfo{pages}{1430--1439}.
%Type = Inproceedings
\bibitem[{Wang et~al.(2018b)Wang, Girshick, Gupta and He}]{wang2018non}
\bibinfo{author}{Wang, X.}, \bibinfo{author}{Girshick, R.},
  \bibinfo{author}{Gupta, A.}, \bibinfo{author}{He, K.}, \bibinfo{year}{2018}b.
\newblock \bibinfo{title}{Non-local neural networks}, in:
  \bibinfo{booktitle}{Conference on Computer Vision and Pattern Recognition
  (CVPR)}, \bibinfo{organization}{IEEE}. pp. \bibinfo{pages}{7794--7803}.
%Type = Inproceedings
\bibitem[{Wang et~al.(2019)Wang, Jabri and Efros}]{wang2019learning}
\bibinfo{author}{Wang, X.}, \bibinfo{author}{Jabri, A.},
  \bibinfo{author}{Efros, A.A.}, \bibinfo{year}{2019}.
\newblock \bibinfo{title}{Learning correspondence from the cycle-consistency of
  time}, in: \bibinfo{booktitle}{Conference on Computer Vision and Pattern
  Recognition (CVPR)}, \bibinfo{organization}{IEEE}. pp.
  \bibinfo{pages}{2566--2576}.
%Type = Inproceedings
\bibitem[{Wu et~al.(2020)Wu, Girshick, He, Feichtenhofer and
  Kr{\"a}henb{\"u}hl}]{wu2019multigrid}
\bibinfo{author}{Wu, C.Y.}, \bibinfo{author}{Girshick, R.},
  \bibinfo{author}{He, K.}, \bibinfo{author}{Feichtenhofer, C.},
  \bibinfo{author}{Kr{\"a}henb{\"u}hl, P.}, \bibinfo{year}{2020}.
\newblock \bibinfo{title}{A multigrid method for efficiently training video
  models}, in: \bibinfo{booktitle}{Conference on Computer Vision and Pattern
  Recognition (CVPR)}, \bibinfo{organization}{IEEE}. pp.
  \bibinfo{pages}{153--162}.
%Type = Inproceedings
\bibitem[{Zhao et~al.(2019)Zhao, Torralba, Torresani and Yan}]{zhao2019hacs}
\bibinfo{author}{Zhao, H.}, \bibinfo{author}{Torralba, A.},
  \bibinfo{author}{Torresani, L.}, \bibinfo{author}{Yan, Z.},
  \bibinfo{year}{2019}.
\newblock \bibinfo{title}{{HACS}: Human action clips and segments dataset for
  recognition and temporal localization}, in: \bibinfo{booktitle}{International
  Conference on Computer Vision (ICCV)}, \bibinfo{organization}{IEEE}. pp.
  \bibinfo{pages}{8668--8678}.
%Type = Inproceedings
\bibitem[{Zhao et~al.(2018)Zhao, Zhang, Liu, Shi, Change~Loy, Lin and
  Jia}]{zhao2018psanet}
\bibinfo{author}{Zhao, H.}, \bibinfo{author}{Zhang, Y.}, \bibinfo{author}{Liu,
  S.}, \bibinfo{author}{Shi, J.}, \bibinfo{author}{Change~Loy, C.},
  \bibinfo{author}{Lin, D.}, \bibinfo{author}{Jia, J.}, \bibinfo{year}{2018}.
\newblock \bibinfo{title}{Psanet: Point-wise spatial attention network for
  scene parsing}, in: \bibinfo{booktitle}{European Conference on Computer
  Vision (ECCV)}, pp. \bibinfo{pages}{267--283}.

\end{thebibliography}

\end{document}